\documentclass[runningheads,a4paper]{llncs}

\usepackage[utf8]{inputenc}
\usepackage{amssymb,amsmath}
\usepackage{graphicx}
\usepackage{url}
\usepackage{tikz, pgfplots}
\pgfplotsset{compat=newest}
\usepackage{url}
\usepackage{paralist}
\usepackage{csquotes}
\MakeOuterQuote{"}

\urldef{\mailsSP}\path|{janno.stuelpnagel, jens.ortmann}@softplant.de|    
\urldef{\mailsMA}\path|{joerg, christian,heiner}@informatik.uni-mannheim.de|    
\newcommand{\keywords}[1]{\par\addvspace\baselineskip
\noindent\keywordname\enspace\ignorespaces#1}
\usepackage{color}
\definecolor{darkred}{RGB}{210,1,1}

\begin{document}

\mainmatter  

\title{Using Abduction in Markov Logic Networks for Root Cause Analysis}

\titlerunning{Using Abduction in Markov Logic Networks for Root Cause Analysis}

%
%
\author{Joerg Schoenfisch\inst{1} \and Janno von Stülpnagel\inst{2} \and Jens Ortmann\inst{2} \and \\ Christian Meilicke\inst{1} \and Heiner Stuckenschmidt\inst{1}}
\authorrunning{Schoenfisch \and von Stülpnagel \and Ortmann \and Meilicke \and Stuckenschmidt}

\institute{Research Group Data and Web Science\\
University of Mannheim, Germany\\
\mailsMA\\
\url{http://dws.informatik.uni-mannheim.de}\\
\and 
Softplant GmbH\\
Agnes-Pockels-Bogen 1\\
80992 Munich, Germany\\
\mailsSP\\
\url{http://www.softplant.de/}
}

%
%

\maketitle

\begin{abstract}
In this paper we propose an approach for calculating the most probable root cause for an observed failure in an IT infrastructure.
Our approach is based on Markov Logic Networks.
While Markov Logic supports a special type of deductive inference, known as maximum a posteriori inference, the computation of the most probable cause requires abductive reasoning.
Abduction aims to find an explanation for a given observation in the light of some background knowledge.
In failure diagnosis, the explanation corresponds to the root cause, the observation corresponds to the failure of a component or service, and the background knowledge corresponds to the dependency graph of the infrastructure extended by potential risks.
We apply the method for abduction proposed by Kate et al. to extend a Markov Logic Network in order to conduct abductive reasoning~\cite{Kate2009}.
We illustrate that our approach is a well suited method for root cause analysis by applying it to a sample scenario.
\end{abstract}

\section{Introduction}
Root cause analysis (RCA) plays an important part in processes for problem solving in many different settings.
Its purpose is to find the underlying source of the observed symptoms of a problem.
Especially in IT infrastructures, short response times to failures (e.g. failing e-mail deliveries, inaccessible websites, or unresponsive accounting systems) are crucial.
Today's IT infrastructures are getting increasingly complex with diverse explicit and implicit dependencies. 
This makes root cause analysis a time intensive task as the cause for a problem might be unclear or the most probable cause might not be the most obvious one.
Therefore, automating the process of root cause analysis and helping an IT administrative to identify the source of a failure or outage as fast as possible is important to achieve a high service level.

In this paper we present our approach to root cause analysis that uses Markov Logic Networks (MLN) and abductive reasoning to enable an engineer to drill down fast on the source of a problem.
Markov Logic Networks provide a formalism that combines logical formulas (to describe dependencies) and probabilities (to express various possible risks) in a single representation.
We focus on abductive reasoning in MLNs and show how it can be used for the purpose of root cause analysis.
To our knowledge, the proposed approach is a novel method to root cause analysis that combines probabilistic and logical aspects in a well-founded framework.

%
%

Within our framework, the IT infrastructure is represented as a logical dependency network that includes various threats to its components.
When a problem occurs, available observations are fed into the system which then generates the Markov Logic Network from the available observations, the given dependency network, and the general background knowledge related to the components of the infrastructure.
Some of these observations might be specified manually, while other observations can be fed into the system automatically via constantly running monitoring software.
These observations are typically incomplete in the sense that not all relevant components are monitored.
Thus, taking the given observations into account, there might still be several explanations for the problem that occurred.

We calculate, via abduction, the most probable cause for the current problem, which is then presented to the user, e.g. the administrator of the IT infrastructure.
The user can then investigate if it is indeed the source of the problem.
This might require to manually check the availability of some component or to analyze a log file.
If the proposed explanation is correct, counter-measures can be introduced immediately.
If the additional observations revealed that the calculated explanation is wrong, the new observations are fed into the system as additional evidence and a better explanation is computed.
This iterative, dialogue-based process is a practicable approach to quickly narrow down on a root cause.

In our approach, we represent the given infrastructure and the possible risks in first-order logic.
This allows us to automatically infer that certain threats are relevant for certain infrastructure components.
Relevant background knowledge can easily be maintained and used to generate the Markov Logic Network.
Moreover, our approach can take into account known probabilities of risks and failures.
These probabilities are derived from expert judgment or statistical data.
Instead of computing multiple candidate explanations, which is possible in purely logic based approaches, we are able to generate the most probable explanation with our approach, while still leveraging the full power of an expressive declarative framework.


This paper is structured as follows. First, we present the theoretical underpinnings of our approach.
In Section~\ref{sec:preliminaries}, we give a brief description to Markov Logic, introduce the general notion of abduction, and explain how abduction can be realized in the context of Markov Logic Networks.
In Section~\ref{sec:rca}, we first present a typical scenario for root cause analysis.
Then, we show how to model this scenario in our framework and describe how to apply abductive reasoning to find the most probable root cause.
Finally, we present a workflow that illustrates how our approach is used in the context of a dialogue-based process.
We show how our approach is related to other works in Section~\ref{sec:related}.
Finally, we discuss the drawbacks and benefits of our approach, and point out some directions for future work in Section~\ref{sec:conclusion}.

\section{Preliminaries}
\label{sec:preliminaries}
This section first describes Markov Logic Networks.
Then we explain abduction and its concrete implementation in the context of Markov Logic Networks.

\subsection{Markov Logic Networks}
\label{sub:markovlogicnetworks}

Markov Logic Networks (MLN) generalize first-order logic and probabilistic graphical models by allowing hard and soft first-order formulas~\cite{Richardson2006}.
Hard formulas are regular first-order formulas, which have to be fulfilled by every interpretation.
An interpretation is also referred to as a possible world.
Soft formulas have weights that support (in case of positive weights) or penalize (in case of negative weights) worlds in which they are satisfied.
The probability of a possible world, one that satisfies all hard formulas, is proportional to the exponential sum of the weights of the soft formulas that are satisfied in that world. 
This corresponds to the common understanding of Markov Networks as log-linear probabilistic model~\cite{Richardson2006}.


An MLN is a template for constructing a Markov Network.
A formula is called a grounded formula if all variables have been replaced by constants. 
Given a set of constants, a Markov Network can be generated from the MLN by computing all possible groundings of the given formulas. 
Due to the closed world assumption, the domain of interest consists of only those entities that are defined by specifying the set of constants.
An atom is a formula that consists of a single predicate.
A possible world corresponds to a set of ground atoms, which is usually a small subset of all possible groundings. 
		
Formally, an MLN $L$ is a set of pairs $\langle F_i,w_i \rangle$, where $F_i$ is a first-order logic formula and  $w_i$ is a real numbered weight \cite{Richardson2006}.
The MLN $L$, combined with a finite set of constants $C = \lbrace c_1,c_2,...c_{|C|} \rbrace$, defines a ground Markov Network $M_{L,C}$ as follows:

\begin{quote}
\begin{enumerate}
	\item $M_{L,C}$ has one binary node for each possible grounding of each predicate in \textit{L}.
		The value of the node is 1 if the grounded atom is true and 0 otherwise.
	\item $M_{L,C}$ contains one feature for each possible grounding of each formula $F_i$ in L.
		The value of this feature is 1 if the formula is true, and 0 otherwise. 
		The weight of the feature is the $w_i$ associated with $F_i$ in L.
\end{enumerate} \cite[p. 113]{Richardson2006}
\end{quote}

Generally, a feature can be any real-valued function of the variables of the network.
In this paper we use binary features, essentially making the value of the function equal to the truth value of the grounded atom.

The description as a log-linear model leads to the following definition for the probability distribution over possible worlds \textit{x} for the Markov Network $M_{L,C}$:
\begin{equation} 
	P(X=x) = \frac{1}{Z} \exp \biggl( \sum_i w_i n_i (x) \biggr)
\end{equation}
where Z is a normalization constant and $n_i(x)$ is the number of true groundings of $F_i$ in $x$.
		
When describing the MLN we use the format $\langle \text{\textit{first-order formula}}, \text{\textit{weight}} \rangle$.
Hard formulas have infinite weights.
If the weight is $+\infty$ the formula must always be true, if the weight is $-\infty$ it must always be false.
A soft formula with weight 0 has equal probabilities for being satisfied in a world or not.

There are two types of inference with Markov Logic: maximum a posteriori (MAP) inference and marginal inference.
MAP inference finds the most probable world given some evidence.
Marginal inference computes the posteriori probability distribution over the values of all variables given some evidence.
We are interested in MAP inference, as we want to determine the world with the most probable explanation for a failure.


\subsection{Abduction in Markov Logic Networks}
\label{sub:abduction}
Abductive reasoning -- or simply \textit{abduction} -- is inference to the best explanation. 
It is applicable to a wide array of fields in which explanations need to be found for given observations, 
for example plan or intent recognition, medical diagnosis, criminology, or, as in our approach, root cause analysis.
According to~\cite{Kate2009}, abduction is usually defined as follows~\cite{Pople1973}:
\begin{description}
\item[Given:] Background knowledge $B$ and a set of observations $O$, both formulated in first-order logic with $O$ being restricted to ground formulae.
\item[Find:] A hypothesis $H$, also a set of logical formulae, such that $B \cup H$ is consistent and $B \cup H \vdash O$.
\end{description}
In other words, find a set of assumptions (a hypothesis) that is consistent with the background knowledge and, combined with it, explains the observation.
It is the opposite of deductive reasoning which infers effects from cause.

The relation between root cause analysis and abductive reasoning is rather straightforward.
In our approach, the background knowledge is the dependency network, respectively the Markov Logic Network to which we transform it.
The dependency graph and Markov Logic Networks both are based on first-order logic as a formalism and thus conveniently are already in the desired logical representation.
The observations, i.e. information about components being available or unavailable, are not part of the model but rather are directly provided as evidence to the MLN.
We then try to prove through abduction that a specific threat -- the most plausible cause -- has occurred.

The inference mechanism in Markov Logic Networks is inherently deductive, not abductive.
Deductive reasoning draws new, logically sound conclusions from given statements.
Kate et al. and Singla et al. \cite{Kate2009,Singla2011a} proposed methods -- Pairwise Constraint (PC) and Hidden Cause (HC) model -- that adapt Markov Logic Networks to automatically perform probabilistic abductive reasoning through its standard deductive reasoning mechanism.
Their method augments the clauses of the MLN to support abductive reasoning as defined above.
In general, the methods first introduce a reverse implication for every logical implication already present in the network.
For example, if there are formulas $p_1 \rightarrow q$, $\ldots$, $p_n \rightarrow q$ in the MLN, the formula $q \rightarrow p_1 \vee \ldots \vee p_n$ is added to the MLN. 

In a second step the model is then extended with mutual exclusivity constraints that bias the inference against choosing multiple explanations.
The reverse implications and the mutual exclusivity clauses are modeled as soft rules and may occasionally be violated, for example, if multiple explanations provide a better proof for the hypothetical root cause than a single explanation.
We follow this basic idea, however, we argue that the mutual exclusivity constraints are not required in the application that we are interested in.

\section{Root Cause Analysis with Markov Logic Networks}
\label{sec:rca}
Root cause analysis is the task of finding the underlying cause of an event.
It is often applied to analyze system failures. System failures are commonly caused by a cascade of events.
The goal of a root cause analysis is finding the original reason for the failure, so that a sustainable solution can be provided~\cite{Rooney2004}.
Root cause analysis typically comprises two phases: the detection of an event and the diagnosis of the event.
In our work, we are concerned with the second phase and assume that a failure has already been detected.

In this section, we first illustrate the infrastructure of our case study.
Afterwards, we show how to model dependencies and risks as a set of first-order formulas.
Then, we explain how we implemented abduction in our Markov Logic Network and show special properties of our settings which simplify the general approach of abductive reasoning.
Finally, we explain how the method is integrated in an iterative process using the example presented at the beginning.

\subsection{Scenario Setting}
\begin{figure}[h!tbp]
	\centering
		\includegraphics[width=\textwidth]{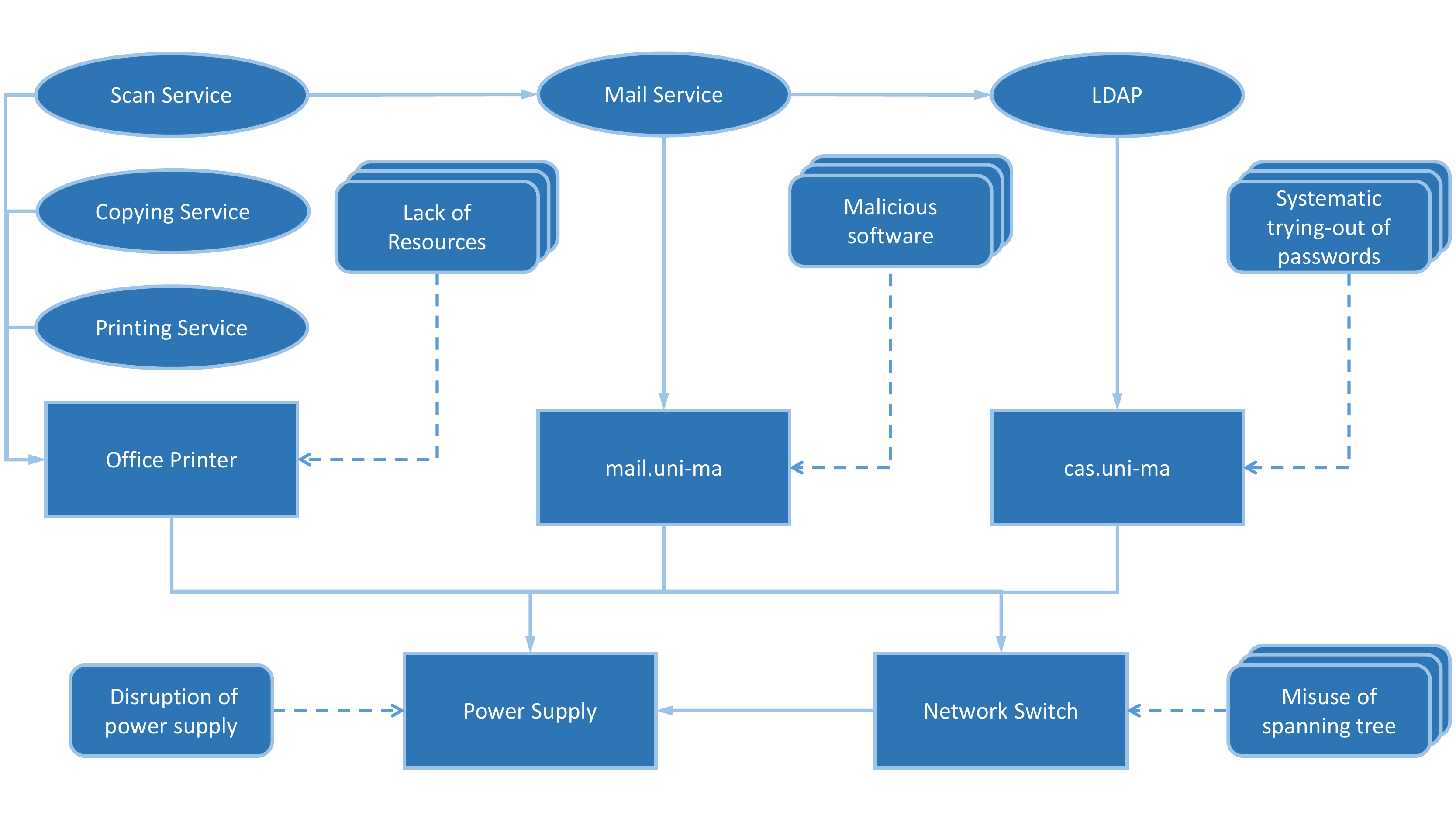}
	\caption{Case Study: Office multifunction printer with risks attached}
	\label{fig:szenario_drucker_risk}
\end{figure}

In the subsequent sections we discuss our approach with the help of an infrastructure shown partially in Figure~\ref{fig:szenario_drucker_risk}. 
This small sample revolving around an office multifunction printer consists of the following components:
\begin{itemize}
	\item The basic dependency for all components is the \textit{Power Supply}.
		The only risk that can affect it is a general outage.
	\item The \textit{Network Switch} connects the other components. 
		It only depends on the power supply; it has multiple risks, e.g. congestion, overheating, or denial-of-service attack, not explicitly depicted in the figure.
	\item The two servers \textit{mail.uni-ma} and \textit{cas.uni-am} each offer one service, i.e. the \textit{Mail Service} and an \textit{LDAP authentication service}.
		The Mail Service uses the LDAP service to authenticate users.
		Both servers have various threats, e.g. malicious software, DOS attacks, overloading, or compromise of the system.
	\item The \textit{Office Printer} offers three services: \textit{Copying}, \textit{Printing}, and \textit{Scanning}.
		It also has various problem sources, e.g. lack of resources or a technical malfunction.
\end{itemize}

The threats we are using in our example are defined in the \textit{IT-Grundschutz Catalogues}~\cite[p. 417ff.]{Informationstechnik2013}:
\begin{itemize}
	\item \textit{Disruption of power supply (T 4.1)}: 
	Short disruption of the power supply, more than 10 ms, or voltage spikes can damage IT devices or produce failures in its operation. 
	\item \textit{Failure of Devices or Systems (T 0.25)}:
	No equipment runs infinitely and a hardware failure in an IT device will happen if it runs long enough. 
	Beyond the damage of the device, the downtime has an effect on the processes that depend on the device or can even damage other devices, e.g. in the case of a cooling system.
	\item \textit{Systematic trying-out of passwords (T 5.18)}:
	An attacker can gain access to a system by discovering the password of the system through systematic trial-and-error.
	\item \textit{Lack of Resources (T 0.27)}: 
	If the given resources (for example bandwidth, disk space or personnel) in an area of the operation are smaller than the current demand a bottleneck occurs. 
	This results in congestion and failure of operation. 
	\item \textit{Malicious software (T 5.23)}: 
	Malicious software tries to execute a process that is unwanted or damaging for the owner of the device that runs the software. 
	This includes viruses, worms and Trojan horses. 
	\item \textit{Misuse of spanning tree (T 5.114)}: 
	An attacker can use Bridge Protocol Data Units (BPDUs) to initialize the recalculation of the switch topology.
	This can be used to disrupt the availability of the network.    
\end{itemize}
The IT-Grundschutz Catalogues are a comprehensive collection of threats and safeguards for various parts of an IT infrastructure\footnote{\url{https://www.bsi.bund.de/EN/Topics/ITGrundschutz/itgrundschutz_node.html}}.
They are created and maintained by the German Federal Office for Information Security\footnote{Bundesamt für Sicherheit in der Informationstechnik (BSI)}, 
and compatible to the ISO 27001 certification\footnote{\url{http://www.iso.org/iso/home/standards/management-standards/iso27001.htm}}.

\subsection{Modeling the Infrastructure}
The foundation of our root cause analysis is the dependency model.
It uses first-order logic to describe various aspects of the IT infrastructure.
Our basic model uses five predicates:
\begin{description}
	\item[specificallyDependsOn(\textit{x,y})] specifies that component $x$ is specifically dependent on component $y$, e.g. the mail service that runs on the mail server.
		This predicate does not allow for any redundancy of $y$.
	\item[genericallyDependsOn(\textit{x,y})] specifies that component $x$ depends on $y$. 
		$y$ may be replaced by some other redundant component.
		An example is a server running on the normal power supply or some uninterruptible power source. 
	\item[redundancy(\textit{x,y})] states that $x$ and $y$ are redundant and $x$ can replace $y$.
	\item[hasRisk(\textit{x,y})] assigns the risk $y$ to component $x$, i.e. $y$ is a threat that endangers the functionality of a component and it can affect $x$.
	\item[unavailable(\textit{x})] designates a component $x$ as unavailable, e.g. offline or not functioning properly.
\end{description}
Formulae~\ref{mln:specific} to~\ref{mln:excl} depict the basic MLN program built from those predicates:
\footnotesize
	\begin{subequations}
\label{fig:mln_program}
		\begin{equation}\label{mln:specific}
			\langle \forall x,y \; (\text{specificallyDependsOn}(x,y) \wedge \text{unavailable}(y) 
				\Rightarrow \text{unavailable}(x)), \infty \rangle
		\end{equation}
		\begin{equation}\label{mln:generic}
		\begin{split}
			\langle \forall x,y \; (\text{genericallyDependsOn}(x,y) &\wedge \text{unavailable}(y)  \\
			\wedge \: \neg \exists z \; (\text{redundancy}(y,z) &\wedge \neg \text{unavailable}(z)) \Rightarrow  \text{unavailable}(x)), \infty \rangle
		\end{split}
		\end{equation}
		\begin{equation}\label{mln:redundancy}
			\langle \forall x,y \; (\text{redundancy}(x,y)
				\Rightarrow \text{redundancy}(x,y)), \infty \rangle
		\end{equation}
		\begin{equation}\label{mln:redundancy2}
			\langle \forall x,y \; (\text{redundancy}(x,y) \wedge \text{redundancy}(y,z)
				\Rightarrow \text{redundancy}(x,z)), \infty \rangle
		\end{equation}
		\begin{equation}\label{mln:risk}
			\langle \forall x,y \; (\text{affectedByRisk}(x,y) 
				\Rightarrow \text{unavailable}(x)),  \infty \rangle
		\end{equation}
		\begin{equation}\label{mln:excl}
			\langle \forall x,y \; \neg(\text{specificallyDependsOn}(x,y) \wedge \text{genericallyDependsOn}(x,y)), \infty \rangle
		\end{equation}
	\end{subequations}
\normalsize

Formula \ref{mln:specific} forbids any world where infrastructure component $x$ is unavailable and infrastructure component $y$ is available, if there is a specific dependency from $x$ to $y$.
Formula~\ref{mln:generic} is similar to Formula \ref{mln:specific}, but for generic dependencies with redundancies.
Provided $x$ is generically dependent on $y$, $y$ is unavailable, and there exists no available component $z$ that is redundant with $y$, then $x$ is also unavailable.
Thus, a component is only available if every specific dependency is available or if at least one redundant component is available for each generic dependency, respectively. 
The symmetry and transitivity of \textit{redundancy} is modeled by Formulae \ref{mln:redundancy} and \ref{mln:redundancy2}.
By adding these two formulas, we ensure that it is not required to specify redundancy for all pairs in both directions.
If we extend an infrastructure with an additional redundant component, we only need to add a single statement instead of specifying the information for all pairs in the group of redundant components.
Formula \ref{mln:risk} enforces that a component $x$ that is affected by the effects of a risk $y$ becomes unavailable.
The predicates \textit{specificallyDependsOn($x,y$)} and \textit{genericallyDependsOn($x,y$)} are mutually exclusive (Formula~\ref{mln:excl}).

The known dependencies, risks, and unavailabilities are modeled as evidence as shown below. Note that these formulas are only two examples for all formulas required to describe the infrastructure depicted in Figure~\ref{fig:szenario_drucker_risk}.
\begin{subequations}\label{fig:mln_evidence}
	\begin{equation}\label{mln:ev_specific}
		\langle \text{specificallyDependsOn}(\text{\textit{MailService}},\text{\textit{mail.uni-ma}}), \infty \rangle
	\end{equation}
	\begin{equation}\label{mln:ev_risk}
		\langle \text{affectedByRisk}(\text{\textit{mail.uni-ma}},\text{\textit{MaliciousSoftware}}), -1.2 \rangle
	\end{equation}
\end{subequations}
Formula \ref{mln:ev_specific} is a hard fact, which states that the \textit{MailService} depends on the server \textit{mail.uni-ma}.
The soft Formula \ref{mln:ev_risk} encodes that \textit{mail.uni-ma} can be affected by \textit{MaliciousSoftware}.
This formula has a negative weight, i.e. has a low probability.
As described before, the dependency relation must hold in every possible world.
The soft formula is not fulfilled in most of the worlds.
In fact, if only this evidence is given, the most probable world does not include it, as it lowers the sum of the weights of all formulas.

Determining the correct weight for the evidence is not trivial.
However, there exist efficient learning algorithms for MLNs \cite{Richardson2006}.

Our basic dependency model only contains relatively simple rules, and only soft formulas in the evidence.
One way in which we extended the MLN program is by adding additional general knowledge about types of components.

For example, we can add information about the failure rate (in the form of a weight) of a specific hard drive model to our MLN program.
\begin{subequations}\label{fig:mln_inherit}
	\begin{equation}\label{mln:in_type}
		\langle \forall x \; (\text{SCSIHardDrive}(x) \Rightarrow \text{affectedByRisk}(\text{\textit{x}},\text{\textit{HeadCrash}})), -1.8 \rangle
	\end{equation}
	\begin{equation}\label{mln:in_ind1}
		\langle \text{SCSIHardDrive}(\textit{DriveInstance1}), \infty \rangle
	\end{equation}
	\begin{equation}\label{mln:in_ind2}
		\langle \text{SCSIHardDrive}(\textit{DriveInstance2}), \infty \rangle
	\end{equation}
\end{subequations}
The hard drive model \textit{SCSIHardDrive} is described as hard drive that has a certain risk of a head crash (\ref{mln:in_type}).
The weight attached to this formula can be derived from available failure rates.
If required we can also add further types related to, e.g. the manufacturer of the drive, since it might be known that drives produced by a certain company have a lower failure rate.
We then can model individual drives as instance of this type (Formulae \ref{mln:in_ind1} and \ref{mln:in_ind2}).
Subsequently, they inherit all the properties, i.e. the weighted risk of a head crash.

This way, we can build hierarchies of components or threats that facilitate populating the model with evidence later.
Another possible usage is to define predicates based on location, e.g. that all components in the same building depend on the same power supply.
Thus, we do not have to specify all information explicitly, but leverage the reasoning capabilities of our approach.

\subsection{Computing Explanations}
We now detail our approach and describe how the Markov Logic Network is constructed and extended, and how we use abductive reasoning for root cause analysis.
The construction from background knowledge and extension for abduction of the Markov Logic Network is only done once and does not have to be changed during the root cause analysis.
According to the method proposed in \cite{Kate2009} we have to add one reverse implication for the Formulae~\ref{mln:specific}, \ref{mln:generic}, and \ref{mln:risk}:
\footnotesize
\begin{equation}
\begin{split}	
	\forall x \; (\text{unavailable}(x) \Rightarrow ( &\exists y \; (\text{specificallyDependsOn}(x,y) \wedge \text{unavailable}(y))) \ \vee \\
	(&\exists y \; (\text{genericallyDependsOn}(x,y) \wedge \text{unavailable}(y)  \\
			&\wedge \: \neg \exists z \; (\text{redundancy}(y,z) \wedge \neg \text{unavailable}(z)))) \ \vee\\
				(&\exists y \; (\text{affectedByRisk}(x,y)))
\end{split}
\label{eq:mln_disjunction}
\end{equation}
\normalsize
Additionally, Kate et al.s' method requires clauses for mutual exclusivity to be added.
The purpose of these clauses is to \textit{"explain away"} multiple causes for an observation and prefer a single one~\cite{Pearl1988}.
The reverse implications as well as the mutual exclusivity clauses are usually modeled as soft clauses.
In general, for each set of reverse implications $P_i$  with the same left-hand side, $(\frac{|P_i|^2 + |P_i|}{2}) \in O(n^2)$ mutual exclusivity clauses are added.

However, different from networks in that general method, our approach exhibits a property that simplifies the additional rules needed for abduction:
All the weights in the evidence are negative -- based on the reasonable assumption that threats and risks only occur rarely, i.e. components are available more than 50\% of the time.
This property allows us to reduce the size of the Markov Logic Network by leaving out the mutual exclusivity clauses completely:
Due to the reverse implication, the MLN solver has to chose one cause to make the clause \textit{true}.
However, as all causes have negative weights and thus every cause set to true is lowering the sum of the weights of a possible world, the solver is already biased against choosing multiple explanations.
This saves us from generating the quadratic number of mutual exclusivity clauses.

\begin{figure}[h!tbp]%
	\includegraphics[width=\columnwidth]{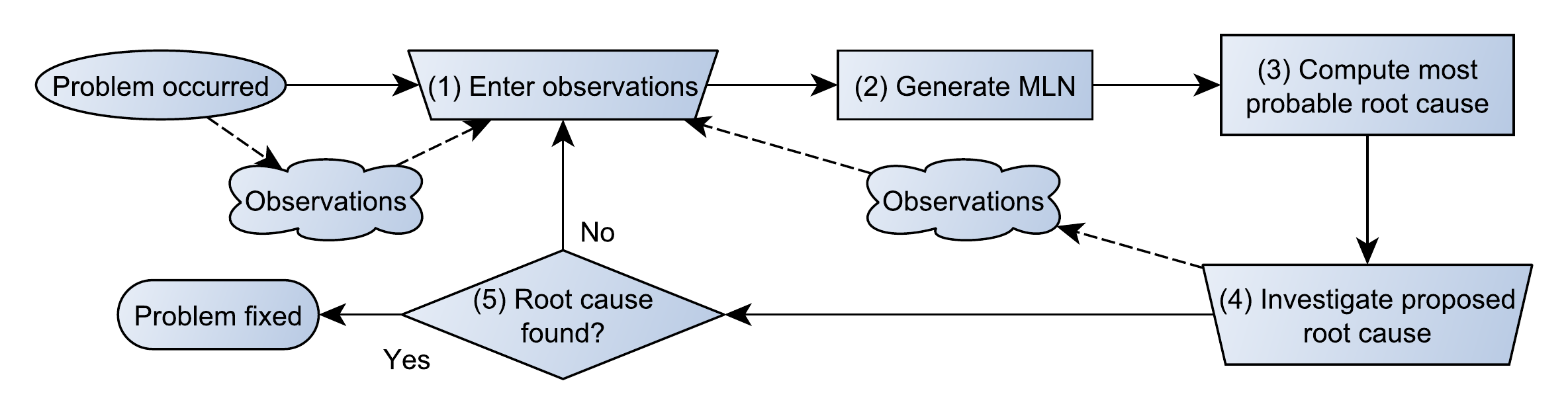}%
	\caption{Process flow for our approach on root cause analysis. Rectangles denote automatic action. Trapezoids require manual interaction by an administrative user.}%
	\label{fig:dialog}%
\end{figure}

After constructing and extending the Markov Logic Network, we can conduct the root cause analysis.
The overall process flow of our approach is depicted in Figure~\ref{fig:dialog}.
The analysis is a dialog-based and iterative process, with interaction between our system and an administrative user.
A fully automatic workflow is desirable, however, not every information can be retrieved directly and sometimes manual investigation of log files or on the status of components is necessary.

In its normal state, without any hard evidence about availabilities or unavailabilities, all components are assumed to be available.
Thus, when calculating the MAP state, it contains all components as available.
When a problem occurs the user is required to provide observations as evidence for the MLN (1).
These observations include any certain information about available and unavailable components. 
For example, the user can enter that printing over the network is not possible, although the network is functional as browsing the internet still works.
This results in hard evidence for the printing service being unavailable and network services and hardware required for internet access being available.

Our approach extends the Markov Logic Network with the new evidence (2) and uses an MLN solver to run MAP inference on it (3).
The calculated MAP state contains the evidence provided by the user (this must be always fulfilled), 
components being unavailable due to a direct or indirect dependency on components observed as not available,
and (at least) one root cause that explains the unavailabilities.
Components which are not affected by specified observations or the calculated root cause are listed as available.

\noindent The root cause fulfills the following properties:
\begin{compactitem}
	\item It explains all unavailabilities in the evidence. 
		This is the case due to the additional reverse implications.
	\item It is not affecting any component stated as available in the evidence. 
		Otherwise a hard rule would be violated.
	\item It is the most probable cause for all the observations given as evidence and the risk probabilities specified as weights.
\end{compactitem}
We make the assumption that all causes are unlikely (they appear less than 50\% of the time).
Thus, their weights are negative.
As the objective of the MAP state is maximizing the sum of all weights, only the most likely cause that explains all observations is included.
A less likely cause has a higher negative weight, causing the sum of the weights to be lower than optimal, and thus getting rejected.
		
Our approach only presents multiple possible root causes, if the sum of their weights is less than the weight of a single possible cause.
If there are two possible root causes with the same weight, only one is presented at random.

The user then has to investigate the presented root cause (5).
If it is the source of the observed problem, the analysis is finished and the cause can be fixed.
Otherwise the process starts over from (1) where the user enters additional observations.
Those new observations can either be gathered while investigating the proposed root cause, 
or, for example, the user can verify the state of components that should also be affected by this cause.

\subsection{Scenario Analysis}
\label{sec:eval}
The following section describes the application of our approach to two different scenarios.
These two scenarios illustrate failures that occurred in our IT infrastructure during the last months.
Together with our system administrators, we modeled our infrastructure, analyzed these scenarios in hindsight, and tested the usefulness of our approach in retrospective.
We used RockIt~\cite{Noessner2013}, a highly optimized and scalable MLN solver, to compute the MAP state.

The first scenario is the one depicted in Figure~\ref{fig:szenario_drucker_risk}, revolving around the malfunction of our office multifunction printer.
The printer offers three services: copying, printing via the network, and scanning to PDF which is then sent to an email address. 
A user reported the printer being broken, as scanning to PDF no longer worked.
To check the proper functioning of the device, the administrator sent a print job and did a photo copy.
Both tests worked successfully.
Sending a test mail from his own account, the administrator also found the mail service working correctly.
Further investigation finally revealed that the root cause of the scanning problem was a suspension of the account the printer used for the LDAP authentication.
However, this cause was only considered after several discussions with two expert administrators involved. 

We applied our approach to this scenario. The MLN was constructed automatically from the background knowledge that we maintained as a set of first-order formulas. We fed in the observations \textit{available(PrintService)}, \textit{available(CopyService)}, and \textit{$\neg$available(ScanService)}
and computed the most probable root cause.
The MAP state that was generated as solution contained the root cause \textit{affectedByRisk(cas.uni-ma, Systematic trying-out of passwords)}.
While we could not definitely decide, in retrospective, if this risk was the underlying reason for the failure of the server \textit{cas.uni-ma}, 
an authentication problem related to \textit{cas.uni-ma} was definitely the cause for the problem.

The second scenario is an outage of our internal subversion server.
It involves more components than the previous scenario and benefits from the iterative approach.
The subversion server is hosted on a virtual machine that is running on a blade server.
Subversion was responding slowly and took long time for many operations.
Neither Subversion nor other processes on the virtual machine showed considerable resource utilization.
Investigating resource usage on the blade server first did not reveal any abnormality.
Later, a user discovered that our external website behaved similarly in performance as the SVN.
This observation was first attributed to a slow internet connection in general, 
but we then discovered that the web server, which was hosted in a different VM but on the same blade server, produced very high network traffic, starving all other services.
A member of our group had released a data set of several gigabytes in size, that was downloaded a few hundred times concurrently.
That lead to congestion on the network interface of the server.
Moving the download to another physical server resolved the problem and the behavior of the subversion server and our website went back to normal.

Analyzing this scenario with our approach, first, we only entered the observation of the unavailability of the SVN service: \textit{$\neg$available(Service\_Subversion)}.
The computed MAP state proposed \textit{affectedByRisk(VM\_Subversion, Overload)} as root cause.
After ruling out this cause by adding \textit{available(VM\_Subversion)} and the observation \textit{$\neg$available(Service\_WebHosting)}, 
the result of the computation was \textit{affectedByRisk(NetworkInterface\_BladeServer, Congestion)} as root cause.
This risk has a high probability for that server which is running various other virtual machines, all hosting services sensitive to a high network load.
The lack of other resources, e.g. CPU or RAM, is modeled as less probable, because all those services are usually not very computational complex or requiring lots of memory.
For this scenario, our approach proposed reasonable root causes which we retrospectively could verify as the reason for the outage.
The manual handling of the incident involved more guesswork by the system administrators and was long winded.



\section{Related Work}
\label{sec:related}
Related work can roughly be divided into two parts: Approaches also conducting root cause analysis, but using a different method;
and approaches using probabilistic frameworks for abductive reasoning, yet not in the context of root cause analysis.

\subsection{Root Cause Analysis}
In previous work, failure diagnosis is conducted using correlation measures.
A specific correlation measure for failure diagnosis is presented in \cite{Marwede2009}.
The approach uses anomalies in the timing of program calls to trace the real root cause of an event.
The anomalies are aggregated to give an anomaly score for each component.
The scores are correlated within their architectural level to determine an anomaly ranking, which  expresses the likelihood that a component is the root cause of a failure. A method for failure diagnosis using decision trees is proposed in \cite{chen2004}.
The decision tree classifies the successful as well as failed requests.
A correlation of paths in the decision tree with occurred failures indicates the node that represents the likely root cause.

In \cite{Zawawy2012} an approach for requirements-driven root cause analysis for failures in software systems is proposed, wherein a Markov Logic Network is used as knowledge repository for diagnostic knowledge.
The approach uses log data as observation information, the  Markov Logic Network is used to deal with uncertainty stemming from incomplete log data.
Their approach differs from ours in several points: they first model the background knowledge as goal trees and only convert it to first-order logic later; the evidence is solely generated from log data; and most importantly they use marginal inference, different to our approach which uses MAP inference. In~\cite{Stulpnagel2014} marginal inference was also used for the purpose of estimating unavailabilities in an IT infrastructure, where the authors referred to problems when marginal inference is applied to very low probabilities usually attached to the occurrence of risks in an IT setting. These problems are based on the use of sampling algorithms for performing marginal inference. Our approach is based on solving an optimization problem, which is not affected negatively by very small probabilities.

The Shrink tool \cite{Kandula2005} uses a Bayesian Network to model the diagnosis problem.
It extends previous work on fault diagnosis with Bayesian Networks \cite{Steinder2002}, by proposing a greedy inference algorithm with polynomial running time.
Furthermore, Shrink is able to handle noise and small inaccuracies in the observations.

\subsection{Applications of Abductive Reasoning}
In \cite{Singla2011a}, Singla et al. extend the approach presented in \cite{Kate2009} and use it in the context of plan and intent recognition.
Instead of adding reverse implication, they introduce a hidden cause for all implications with the same left-hand side. In general, this reduces the size of the MLN and subsequently increases performance.
However, as detailed above, for our approach the mutual exclusivity clauses are not needed anyway.
Nonetheless, if more probable events have to be included in the evidence, their optimization can also be included in our approach. 


Most other approaches to abductive reasoning either use first-order logic to calculate a minimal set of assumptions sufficient to explain the hypothesis \cite{Poole1987,Stickel1991,Ng1991,Kakas1992}, or Bayesian Networks to compute the posterior probability of alternative explanations given the observations \cite{Pearl1988}.
The former approaches are not able to estimate the likelihood of alternative explanations, as they do not support uncertainty in the background knowledge or evidence.
Bayesian Networks, on the other hand, are designed to handle uncertainty.
However, as they are propositional in nature, they cannot handle structured knowledge involving relations amongst multiple entities directly \cite{Kate2009}.

Bayesian Abductive Logic Programs (BALP)~\cite{Raghavan2010} are another approach that combines first-order logic and probabilistic graphical models.
The main difference to MLNs is that BALPs are based on Bayesian Networks, which are directed.
Undirected relations, like the symmetry of redundancy, are thus more complex to model. In \cite{Inoue2011}, Inoue et al. describe a system that uses integer linear programming (ILP) for weighted abduction.
They outperform a state-of-the-art abductive engine\footnote{Mini-\textsc{Tacitus}: \url{http://www.rutumulkar.com/tacitus.html}}.
The MLN solver we use also transforms the problem internally to an ILP, which is one of the reasons for its good runtime performance.

%

\section{Discussion and Conclusion}
\label{sec:conclusion}

We presented our approach of applying abductive reasoning using Markov Logic Networks to compute the most probable root cause for a failure in an IT infrastructure.
Our approach models the infrastructure with the help of first-order logic.
In particular, we formulated the dependencies of the network as hard formulas.
Moreover, we added weighted soft formulas to model the probability of risks that might result in the failure of components and services.
We defined these risks in accordance to the taxonomy of the IT Grundschutz Catalogues.
Furthermore, we have argued how the expressiveness of first-order logic can be used to model general, reusable knowledge concerning risks and IT components.
Our approach uses the same formalism for both knowledge presentation and abductive reasoning.
Thus, all relevant information is readily available to compute the most probable root cause once an incident occurs.
To the best of our knowledge, there exists no other approach that combines uncertainty and logical abductive reasoning to solve the problem of root cause analysis. 


We conducted an evaluation of our approach by analyzing two failures that happened in the infrastructure of our research group.
In both cases we were able to determine a root cause (respectively, a sequence of probable root causes) that turned out to be helpful for a system administrator to resolve the problem.
Our approach is especially useful when the reasons for the failure are not obvious to the administrator that is in charge of resolving the problem.
Thus, our approach will be more beneficial in IT infrastructures, where competences are scattered over the members of different organizational units. 

We did not conduct an evaluation of the scalability of our approach.
However, the MLN solver RockIt, which we used, was extensively tested in other complex, large scale settings and showed good performance for MLNs with several hundred formulas and tens of thousands of facts in the evidence~\cite{Noessner2013,Noessner2014}. 

So far, we have not taken all risks of the Grundschutz Catalogues into account.
Instead, we have focused on a subset relevant for the infrastructure we modeled.
To apply our approach to an arbitrary IT infrastructure, we have to create a complete translation of the catalogues to our logical representation.
Further, we plan to implement a user interface that presents the computed root cause in a comprehensive way and allows to specify observations without the need for understanding the underlying formalism.
Currently, the user has to model all observations as logical formulas.
Once we set up such an interface, we are able to study performance, acceptance, and benefits of the approach in a field study where the complete system is used in the daily work of a computer center.


\bibliographystyle{splncs}
\bibliography{risikomanagement-2015_CAiSE}

\end{document}